\documentclass[conference]{IEEEtran}
\IEEEoverridecommandlockouts
\usepackage{cite}
\usepackage{amsmath,amssymb,amsfonts}
\usepackage{algorithmic}
\usepackage{graphicx}
\usepackage{textcomp}
\usepackage{xcolor}
\usepackage{booktabs} 
\usepackage{multirow}
\def\BibTeX{{\rm B\kern-.05em{\sc i\kern-.025em b}\kern-.08em
    T\kern-.1667em\lower.7ex\hbox{E}\kern-.125emX}}
\begin{document}

\title{Comparative Analysis of Open-Source Language Models in Summarizing Medical Text Data\\
}
\author{
\IEEEauthorblockN{Yuhao Chen\IEEEauthorrefmark{1}\IEEEauthorrefmark{3}, Zhimu Wang\IEEEauthorrefmark{1}\IEEEauthorrefmark{4}, Bo Wen\IEEEauthorrefmark{2}\IEEEauthorrefmark{6}, Farhana Zulkernine\IEEEauthorrefmark{1}\IEEEauthorrefmark{5}}

\IEEEauthorblockA{\IEEEauthorrefmark{1}School of Computing, \IEEEauthorrefmark{2}Digital Health Department, \\}
\IEEEauthorblockA{\IEEEauthorrefmark{1}Queen's University, Kingston, Ontario, Canada
}
\IEEEauthorblockA{\IEEEauthorrefmark{2}IBM Research,Yorktown Heights, NY, USA
}
\IEEEauthorblockA{ \{yuhao.chen\IEEEauthorrefmark{3},19zw28\IEEEauthorrefmark{4}, farhana.zulkernine\IEEEauthorrefmark{5}\}@queensu.ca \ bwen@us.ibm.com\IEEEauthorrefmark{6}
}
}


\maketitle

\begin{abstract}
Unstructured text in medical notes and dialogues contains rich information. Recent advancements in Large Language Models (LLMs) have demonstrated superior performance in question answering and summarization tasks on unstructured text data, outperforming traditional text analysis approaches. However, there is a lack of scientific studies in the literature that methodically evaluate and report on the performance of different LLMs, specifically for domain-specific data such as medical chart notes. We propose an evaluation approach to analyze the performance of open-source LLMs such as Llama2 and Mistral for medical summarization tasks, using GPT-4 as an assessor. Our innovative approach to quantitative evaluation of LLMs can enable quality control, support the selection of effective LLMs for specific tasks, and advance knowledge discovery in digital health.
\end{abstract}

\begin{IEEEkeywords}
Biomedical summarization, Large Language Model, Generative Model
\end{IEEEkeywords}

\section{Introduction}

An increasing number of researchers have recognized the significance of evaluating the alignment between human intent and generated responses by LLMs \cite{wang2023large}. Conventional similarity-based evaluation metrics, such as ROUGE \cite{lin-2004-rouge} and BERTSCORE \cite{zhang2019bertscore}, are not able to capture this alignment accurately \cite{zheng2023judging}. With the success of the advanced GPT-4 model \cite{Achiam2023GPT4TR}, there has been a growing interest in developing human-free automatic evaluation frameworks that leverage GPT-4's capabilities. Rather than deploying GPT-4 to grade each summary and generate evaluation scores, we employ it to select the better summary from two candidates.

The primary objectives of this study are to: a) establish a pipeline for each LLM to perform biomedical summarization; b) select a suitable prompt for each specific summarization task; and c) design and implement an evaluation framework using GPT-4 as the evaluator. All data used in this study are from publicly available sources.

We develop a two-stage pipeline for evaluating two renowned LLMs, Llama2 \cite{llama2} and Mistral \cite{jiang2023mistral} using uniform prompts in three distinct medical summarization tasks: consumer health question summarization, biomedical query-based summarization, and dialogue summarization. We use multiple datasets, namely MEDIQA-QS \cite{ben-abacha-etal-2021-overview}, MeQSum\cite{ben-abacha-demner-fushman-2019-summarization}, MEDIQA-ANS \cite{savery2020question}, MEDIQA-MAS \cite{ben-abacha-etal-2021-overview}, and iCliniq \cite{zeng-etal-2020-meddialog} for the tasks. Subsequently, GPT-4 serves as an assessor to determine which LLM produces a more effective response, accompanied by an explanation to compare each selected open-source LLM and GPT-3.5.

Our approach addresses the limitations of existing methods by systematically evaluating and comparing LLMs for domain-specific data. By employing GPT-4 as an assessor focusing on coherence, consistency, fluency, and relevance, we contribute to the selection of the most suitable LLM for specific tasks in digital health.

\section{Related Work}
\subsection{Generative Large Language Models}
GPT-3 \cite{instructGPT} acquired the ability of ``in-context learning" after incorporating a substantial amount of additional training data and increasing the parameters by over 100 times compared to GPT-2 \cite{GPT2}. The success of chatbot application, ChatGPT\cite{Achiam2023GPT4TR}\cite{instructGPT}, has led to explosive growth in popularity of decoder-only LLM. Following the success of the GPT-4 model, many new open-source generative LLMs with similar capabilities were published, such as Llama \cite{llama2}, and Mistral \cite{jiang2023mistral}.

\subsection{Evaluating LLMs on Biomedical Summarization}
Jahan et al. \cite{jahan-etal-2023-evaluation} conducted a comparative analysis of the State-Of-The-Art (SOTA) fine-tuned based LLMs BioBART \cite{biobart} and ChatGPT (gpt-3.5-turbo) on abstractive summarization datasets. The findings revealed that ChatGPT with zero-shot setting exhibited superior performance when BioBART lacked sufficient in-domain training data, highlighting ChatGPT's robustness where sufficient training data is not available.

\section{Methodology}
\subsection{Text Summarization}
In this pilot study, we limit our comparative analysis to two open-source models: Llama2-70B and Mistral-7B. GPT-3.5 is the pioneering LLM that is intergrated in ChatGPT, which has been used in comparative analysis in subsequent LLMs research \cite{llama2}. Therefore, GPT-3.5 is selected as the target model in our adversarial assessment strategy (Section \ref{sec:adversarial_strategy}). We follow the prompt design methodology outlined in Jahan et al.'s \cite{jahan-etal-2023-evaluation} work.

\noindent\textit{\textbf{1) LLMs}}
\hfill\\
\indent\textit{Llama2-70B-chat-hf}: Llama2 \cite{llama2} is a collection of transformer decoder-based LLMs, ranging from 7 billion to 70 billion parameters. We use its official optimized iteration, Llama2-70B-chat-hf, for the experiment.

\textit{Mistral-7B-Instruct-v0.1}: Mistral \cite{jiang2023mistral} is a transformer decoder-based LLM with 7 billion parameters, achieving superior performance and efficiency among small-scale LLMs. We select the officially fine-tuned version on conversational datasets, Mistral-7B-Instruct-v0.1.

\textit{GPT-3.5-turbo}: GPT-3.5-turbo \cite{instructGPT} is another transformer decoder-based LLM with undisclosed architecture and number of parameters.
 \begin{figure*}[!h]
     \centering
     \includegraphics[width=1\linewidth]{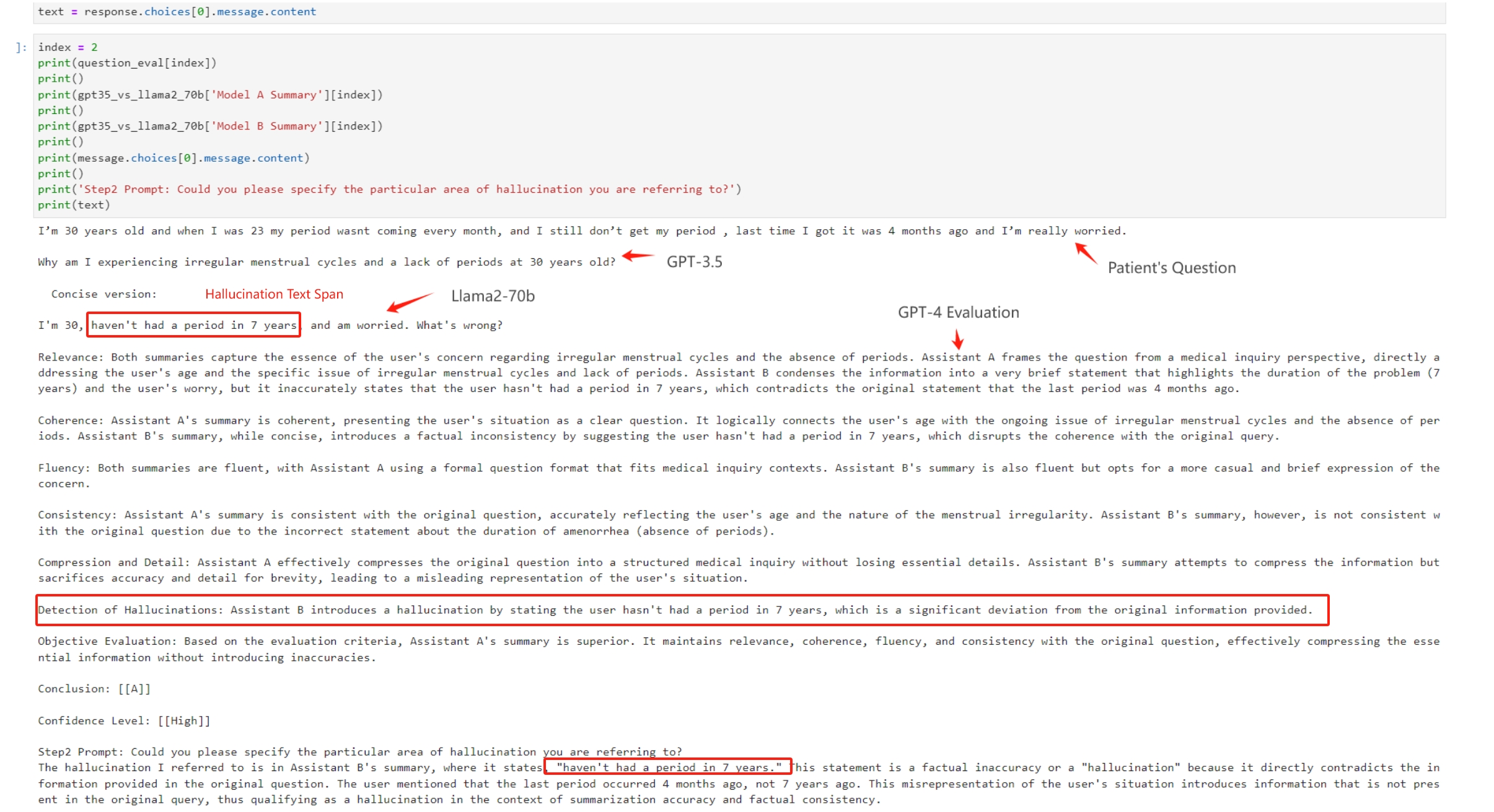}
     \caption{Assessment Example of GPT-4 for LLMs}
     \label{fig:assessment example}
 \end{figure*}
\noindent\textit{\textbf{2) Datasets and Tasks}}
\hfill\\
\indent\textit{Consumer Health Question Summarization}: The objective is to identify the optimal model in summarizing long and complex consumer health questions. Given a customer query $Q$, we combine it with the task-specific instruction $T_q$ to create a prompt $P_q$ that is fed into the model. The methodology for constructing this prompt is shown in Eq. \ref{question_summarization_eq}, with the symbol ``$;$" denoting concatenation. We choose MEDIQA-QS\cite{ben-abacha-etal-2021-overview} and MeQSum \cite{ben-abacha-demner-fushman-2019-summarization} datasets for this task.

\textit{MEDIQA-QS} comprises 150 de-identified consumer health questions sourced from the U.S. National Library of Medicine (NLM), with gold-standard summaries crafted by medical experts. The dataset is divided into validation and test sets, where the validation set encompasses 50 questions, and the test set contains 100 questions. In this study, only the test set is used for evaluation.
  
\textit{MeQSum} is designed for medical problem summarization tasks. It contains 1,000 consumer health questions summarized to facilitate natural language processing (NLP) research specifically for the medical field. The dataset is partitioned into two segments: 500 questions for validation and 500 for the test set. We only use test set for evaluation \cite{jahan-etal-2023-evaluation}\cite{biobart}.

\begin{equation} \label{question_summarization_eq}
\text{$P_q = [T_q;Q_q]$}
\end{equation}

\textit{Biomedical Query-based Summarization}: This task enables models to aggregate and summarize multiple relevant responses to a medical query. For this task, we concatenate the task instruction $T_a$, a medical query $Q_a$, and a referenced document $D_a$ to formulate the prompt $P_a$ for the model as shown in Eq. \ref{answer_summarization_eq}. We choose MEDIQA-ANS \cite{savery2020question} and MEDIQA-MAS \cite{ben-abacha-etal-2021-overview} datasets for this task.

\textit{MEDIQA-MAS} consists of collections of responses to specific health-related queries, along with expert created summaries that integrate information from multiple answers into a single coherent summary. For evaluation, we use the test set, which contains 80 medical questions.
  
\textit{MEDIQA-ANS} contains 156 health questions asked by consumers, their corresponding answers, and expert-created summaries. The dataset is sourced from reliable organizations such as MedlinePlus and derived from questions submitted to the National Library of Medicine's CHiQA system. In our study, we only use "single-document abstractive" type data, which contains 552 data samples for evaluation.

\begin{equation} \label{answer_summarization_eq}
\text{$P_a = [T_a;Q_a;D_a]$}
\end{equation}

\textit{Dialog Summarization}: The goal is to produce a short summary describing a patient's medical conditions from a dialogue between the patient and the doctor. Given a conversation dialogue $Dialog_d$, we combine it with the task instruction $T_d$ to create the prompt $P_d$ as shown in Eq. \ref{dialog_summarization}. We choose the iCliniq \cite{zeng-etal-2020-meddialog} dataset for this task.
  
\textit{iCliniq} is an extensive collection capturing real-life anonymized medical consultations between patients and physicians. It covers a wide range of medical inquiries and responses, demonstrating a doctor's approach to various health problems. In this study, we randomly select 200 data samples for evaluation.

\begin{equation} \label{dialog_summarization}
\text{$P_d = [T_d;Dialog_d]$}
\end{equation}

\begin{table*}[h]
\caption{Comparative Performance of Large Language Models Versus GPT-3.5 Baseline with GPT-4 as the evaluator} \label{tbl:result}
\centering
\tabcolsep=0.09cm
\begin{tabular}{@{}cccccccccccccccc@{}}
\toprule
\multirow{3}{*}{\textbf{Model A Candidates}}             & \multicolumn{3}{c|}{\textbf{MEDIQA-QS}}       & \multicolumn{3}{c|}{\textbf{MeQSum}}          & \multicolumn{3}{c|}{\textbf{MEDIQA-ANS}}      & \multicolumn{3}{c|}{\textbf{MEDIQA-MAS}}      & \multicolumn{3}{c}{\textbf{iCliniq}} \\ \cmidrule(l){2-16} 
                                            & \multicolumn{15}{c}{\textbf{Win Rate}}                                                                                                                                                                                                        \\ \cmidrule(l){2-16} 
                                            &\textbf{Model A} & \textbf{GPT-3.5} & \multicolumn{1}{c|}{\textbf{Tie}}  & \textbf{Model A} & \textbf{GPT-3.5} & \multicolumn{1}{c|}{\textbf{Tie}}  & \textbf{Model A} & GPT-3.5 & \multicolumn{1}{c|}{\textbf{Tie}}  & \textbf{Model A} & \textbf{GPT-3.5} & \multicolumn{1}{c|}{\textbf{Tie}}  & \textbf{Model A}     & \textbf{GPT-3.5}     & \textbf{Tie}      \\ \midrule
\multicolumn{1}{l}{Llama2-70b} & 43\%    & 17\%    & \multicolumn{1}{c|}{40\%} & 42\%    & 18\%    & \multicolumn{1}{c|}{40\%} & 43\%    & 22\%    & \multicolumn{1}{c|}{35\%} & 40\%    & 38\%    & \multicolumn{1}{c|}{22\%} & 44\%        & 16\%        & 40\%     \\
\multicolumn{1}{l}{Mistral-7b}    & 19\%    & 36\%    & \multicolumn{1}{c|}{45\%} & 14\%    & 51\%    & \multicolumn{1}{c|}{35\%} & 40\%    & 24\%    & \multicolumn{1}{c|}{36\%} & 31\%    & 38\%    & \multicolumn{1}{c|}{31\%} & 23\%        & 37\%        & 40\%     \\ \bottomrule
\end{tabular}
\end{table*}

\subsection{Proposed Adversarial Assessment Strategy} \label{sec:adversarial_strategy}
Rather than employing GPT-4 to grade the summaries of each LLM sequentially, we instruct GPT-4 to perform a pairwise comparison \cite{zheng2023judging} between the outputs of two LLMs to determine the superior response. To enhance efficiency, we introduce a novel adversarial assessment strategy that eliminates the need to evaluate all possible pairs by selecting a target LLM and comparing all other LLMs against this target. This methodology reduces the number of evaluations per data sample to $n-1$, where $n$ represents the total number of models under evaluation.

To avoid position bias, we apply the swapping positions approach \cite{zheng2023judging}, employing GPT-4 to do the evaluation twice by swapping the order of two answers. A declaration of victory is made only when an answer is preferred in both orders; otherwise, a tie is declared.

\subsection{Prompt Design}
Our prompt design draws inspiration from the work of Zeng et al. \cite{zheng2023judging}. We enhance the prompt to ensure that GPT-4's evaluation focuses on four key dimensions commonly used by humans: coherence, consistency, fluency, and relevance. 

\subsection{Results}
Table \ref{tbl:result} shows the results of using GPT-4 as an evaluator. Overall, the GPT-4 evaluator demonstrates a preference for summaries generated by Llama2-70b, achieving a rate exceeding 40\%, making it the winner among all evaluated LLMs in all five benchmarks. Mistral-7B only surpasses GPT-3.5 in the MEDIQA-ANS dataset with a 40\% win rate, but it is still lower than Llama2-70b's 43\% win rate.

\section{Discussion and Future Work}
In this pilot study, we acknowledge the limitation of not incorporating additional prompt engineering, relying on existing work for summarization prompts. Future work could benefit from exploring customized prompt designs that might better capture the nuances of medical summarization tasks and enhance model performance.

During the analysis of the GPT-4 evaluator outcomes, we found that GPT-4 assessor demonstrated robust hallucination detection capabilities, as shown in Fig. \ref{fig:assessment example}. However, we also noticed that GPT-4 lacks sensitivity in detecting the length of summaries. Addressing this limitation through prompt engineering or alternative evaluation strategies could improve the accuracy of model comparisons and the overall utility of the evaluation framework. 

While the study proposes to address the alignment between human and GPT-4 evaluator in future work, incorporating an initial investigation into this aspect could strengthen the current findings. Understanding how well the GPT-4 assessor's judgments align with human experts could provide valuable insights into the reliability and applicability of the proposed evaluation framework.

The results of this study have implications for the digital health domain, as they demonstrate the potential of using open-source LLMs and GPT-4 as an assessor for medical summarization tasks. By systematically evaluating and comparing LLMs, our approach can support the selection of the most suitable model for specific tasks, such as summarizing consumer health questions, aggregating medical query responses, and generating dialogue summaries. This can ultimately lead to improved knowledge discovery and decision-making in digital health applications.

We are conducting a human-blind review to validate the GPT-4 assessor's performance with an optimized evaluation prompt which is not presented in this paper. In this evaluation, a human reviewer must assess which summary is better among the pairs generated by different LLMs. The human-blind review is still ongoing. Currently, we have completed evaluations of 200 summary pairs. GPT-4 has achieved an average accuracy and F1 score exceeding 70\%, establishing a solid baseline for the effectiveness of using an LLM as a judge. Moreover, GPT-4 completed 50-question summarization and evaluation tasks in approximately 40 minutes, significantly faster than human annotators, who needed over 2 hours for the same number of questions.

Future research directions include expanding the range of LLMs evaluated, incorporating additional medical summarization datasets, and exploring the impact of domain-specific fine-tuning on model performance. Additionally, investigating the ethical implications of using LLMs in healthcare, such as data privacy, bias, and transparency, is crucial to ensure the responsible deployment of these models in real-world digital health applications.

\section{Conclusion}
This study demonstrates a solution to address the critical need for systematic evaluation and comparison of LLMs in the context of medical text summarization tasks. Our proposed evaluation approach, employing GPT-4 as an assessor, can enable quality control, support the selection of effective LLMs for specific tasks, and advance knowledge discovery in digital health.

\section{ACKNOWLEDGMENT}

We are grateful for the support from NSERC Discovery, CFI, RTI, and IBM for their funding of this research.


\bibliographystyle{IEEEtran}
\bibliography{conference_101719}

\begin{thebibliography}{10}
\providecommand{\url}[1]{#1}
\csname url@samestyle\endcsname
\providecommand{\newblock}{\relax}
\providecommand{\bibinfo}[2]{#2}
\providecommand{\BIBentrySTDinterwordspacing}{\spaceskip=0pt\relax}
\providecommand{\BIBentryALTinterwordstretchfactor}{4}
\providecommand{\BIBentryALTinterwordspacing}{\spaceskip=\fontdimen2\font plus
\BIBentryALTinterwordstretchfactor\fontdimen3\font minus \fontdimen4\font\relax}
\providecommand{\BIBforeignlanguage}[2]{{%
\expandafter\ifx\csname l@#1\endcsname\relax
\typeout{** WARNING: IEEEtran.bst: No hyphenation pattern has been}%
\typeout{** loaded for the language `#1'. Using the pattern for}%
\typeout{** the default language instead.}%
\else
\language=\csname l@#1\endcsname
\fi
#2}}
\providecommand{\BIBdecl}{\relax}
\BIBdecl

\bibitem{wang2023large}
P.~Wang, L.~Li, L.~Chen, D.~Zhu, B.~Lin, Y.~Cao, Q.~Liu, T.~Liu, and Z.~Sui, ``Large language models are not fair evaluators,'' \emph{arXiv preprint arXiv:2305.17926}, 2023.

\bibitem{lin-2004-rouge}
C.-Y. Lin, ``{ROUGE}: A package for automatic evaluation of summaries,'' in \emph{Text Summarization Branches Out}.\hskip 1em plus 0.5em minus 0.4em\relax Barcelona, Spain: Association for Computational Linguistics, Jul. 2004, pp. 74--81.

\bibitem{zhang2019bertscore}
T.~Zhang, V.~Kishore, F.~Wu, K.~Q. Weinberger, and Y.~Artzi, ``Bertscore: Evaluating text generation with bert,'' in \emph{International Conference on Learning Representations}, 2019.

\bibitem{zheng2023judging}
L.~Zheng, W.-L. Chiang, Y.~Sheng, S.~Zhuang, Z.~Wu, Y.~Zhuang, Z.~Lin, Z.~Li, D.~Li, E.~Xing, H.~Zhang, J.~E. Gonzalez, and I.~Stoica, ``Judging {LLM}-as-a-judge with {MT}-bench and chatbot arena,'' in \emph{Thirty-seventh Conference on Neural Information Processing Systems Datasets and Benchmarks Track}, 2023.

\bibitem{Achiam2023GPT4TR}
\BIBentryALTinterwordspacing
OpenAI, ``Gpt-4 technical report,'' 2023. [Online]. Available: \url{https://api.semanticscholar.org/CorpusID:257532815}
\BIBentrySTDinterwordspacing

\bibitem{llama2}
H.~Touvron, L.~Martin, K.~Stone, P.~Albert, A.~Almahairi, Y.~Babaei, N.~Bashlykov, S.~Batra, P.~Bhargava, S.~Bhosale \emph{et~al.}, ``Llama 2: Open foundation and fine-tuned chat models,'' \emph{arXiv preprint arXiv:2307.09288}, 2023.

\bibitem{jiang2023mistral}
A.~Q. Jiang, A.~Sablayrolles, A.~Mensch, C.~Bamford, D.~S. Chaplot, D.~d.~l. Casas, F.~Bressand, G.~Lengyel, G.~Lample, L.~Saulnier \emph{et~al.}, ``Mistral 7b,'' \emph{arXiv preprint arXiv:2310.06825}, 2023.

\bibitem{ben-abacha-etal-2021-overview}
A.~Ben~Abacha, Y.~Mrabet, Y.~Zhang, C.~Shivade, C.~Langlotz, and D.~Demner-Fushman, ``Overview of the {MEDIQA} 2021 shared task on summarization in the medical domain,'' in \emph{Proceedings of the 20th Workshop on Biomedical Language Processing}.\hskip 1em plus 0.5em minus 0.4em\relax Online: Association for Computational Linguistics, Jun. 2021, pp. 74--85.

\bibitem{ben-abacha-demner-fushman-2019-summarization}
A.~Ben~Abacha and D.~Demner-Fushman, ``On the summarization of consumer health questions,'' in \emph{Proceedings of the 57th Annual Meeting of the Association for Computational Linguistics}, A.~Korhonen, D.~Traum, and L.~M{\`a}rquez, Eds.\hskip 1em plus 0.5em minus 0.4em\relax Florence, Italy: Association for Computational Linguistics, Jul. 2019, pp. 2228--2234.

\bibitem{savery2020question}
M.~Savery, A.~B. Abacha, S.~Gayen, and D.~Demner-Fushman, ``Question-driven summarization of answers to consumer health questions,'' \emph{Scientific Data}, vol.~7, no.~1, p. 322, 2020.

\bibitem{zeng-etal-2020-meddialog}
G.~Zeng, W.~Yang, Z.~Ju, Y.~Yang, S.~Wang, R.~Zhang, M.~Zhou, J.~Zeng, X.~Dong, R.~Zhang, H.~Fang, P.~Zhu, S.~Chen, and P.~Xie, ``{M}ed{D}ialog: Large-scale medical dialogue datasets,'' in \emph{Proceedings of the 2020 Conference on Empirical Methods in Natural Language Processing (EMNLP)}, B.~Webber, T.~Cohn, Y.~He, and Y.~Liu, Eds.\hskip 1em plus 0.5em minus 0.4em\relax Online: Association for Computational Linguistics, Nov. 2020, pp. 9241--9250.

\bibitem{GPT3}
T.~Brown, B.~Mann, N.~Ryder, M.~Subbiah, J.~D. Kaplan, P.~Dhariwal, A.~Neelakantan, P.~Shyam, G.~Sastry, A.~Askell \emph{et~al.}, ``Language models are few-shot learners,'' \emph{Advances in neural information processing systems}, vol.~33, pp. 1877--1901, 2020.

\bibitem{GPT2}
A.~Radford, J.~Wu, R.~Child, D.~Luan, D.~Amodei, I.~Sutskever \emph{et~al.}, ``Language models are unsupervised multitask learners.''

\bibitem{instructGPT}
L.~Ouyang, J.~Wu, X.~Jiang, D.~Almeida, C.~Wainwright, P.~Mishkin, C.~Zhang, S.~Agarwal, K.~Slama, A.~Ray \emph{et~al.}, ``Training language models to follow instructions with human feedback,'' \emph{Advances in Neural Information Processing Systems}, vol.~35, pp. 27\,730--27\,744, 2022.

\bibitem{llama1}
H.~Touvron, T.~Lavril, G.~Izacard, X.~Martinet, M.-A. Lachaux, T.~Lacroix, B.~Rozi{\`e}re, N.~Goyal, E.~Hambro, F.~Azhar \emph{et~al.}, ``Llama: Open and efficient foundation language models,'' \emph{arXiv e-prints}, pp. arXiv--2302, 2023.

\bibitem{almazrouei2023falcon}
E.~Almazrouei, H.~Alobeidli, A.~Alshamsi, A.~Cappelli, R.~Cojocaru, M.~Debbah, {\'E}.~Goffinet, D.~Hesslow, J.~Launay, Q.~Malartic \emph{et~al.}, ``The falcon series of open language models,'' \emph{arXiv preprint arXiv:2311.16867}, 2023.

\bibitem{jahan-etal-2023-evaluation}
I.~Jahan, M.~T.~R. Laskar, C.~Peng, and J.~Huang, ``Evaluation of {C}hat{GPT} on biomedical tasks: A zero-shot comparison with fine-tuned generative transformers,'' in \emph{The 22nd Workshop on Biomedical Natural Language Processing and BioNLP Shared Tasks}.\hskip 1em plus 0.5em minus 0.4em\relax Toronto, Canada: Association for Computational Linguistics, Jul. 2023, pp. 326--336.

\bibitem{biobart}
H.~Yuan, Z.~Yuan, R.~Gan, J.~Zhang, Y.~Xie, and S.~Yu, ``{B}io{BART}: Pretraining and evaluation of a biomedical generative language model,'' in \emph{Proceedings of the 21st Workshop on Biomedical Language Processing}.\hskip 1em plus 0.5em minus 0.4em\relax Dublin, Ireland: Association for Computational Linguistics, May 2022, pp. 97--109.

\end{thebibliography}

\end{document}